\documentclass[11pt]{article}

\usepackage[preprint]{acl}

\usepackage{times}
\usepackage{latexsym}
\usepackage{amsmath}
\usepackage{xcolor}
\usepackage{tabularx}
\definecolor{cRole}{HTML}{1A73E8}    
\definecolor{cLogic}{HTML}{B8860B}   
\definecolor{cReq}{HTML}{008080}     
\definecolor{cFormat}{HTML}{800080}  
\definecolor{cConstraints}{HTML}{008080} 

\usepackage[T1]{fontenc}

\usepackage[utf8]{inputenc}

\usepackage{microtype}

\usepackage{inconsolata}

\usepackage{graphicx}
\usepackage{booktabs}

\title{Locomo-Plus: Beyond-Factual Cognitive Memory \\Evaluation Framework for LLM Agents}

\author{
\textbf{Yifei Li}$^{1}$ \quad \textbf{Weidong Guo}$^{2}$ \quad \textbf{Lingling Zhang}$^{1}$ \quad \textbf{Rongman Xu}$^{1}$ \quad \textbf{Muye Huang}$^{1}$\\
\textbf{Hui Liu}$^{2}$ \quad \textbf{Lijiao Xu}$^{2}$ \quad \textbf{Yu Xu}$^{2}$ \quad \textbf{Jun Liu}$^{1}$\\[0.3ex]
$^{1}$Xi'an Jiaotong University \quad $^{2}$Tencent\\
\texttt{yifeilee@stu.xjtu.edu.cn}
}

\begin{document}
\maketitle
\begin{abstract}
Long-term conversational memory is a core capability for LLM-based
dialogue systems, yet existing benchmarks and evaluation protocols
primarily focus on surface-level factual recall.
In realistic interactions, appropriate responses often depend on
implicit constraints such as user state, goals, or values that are not
explicitly queried later.
To evaluate this setting, we introduce \textbf{LoCoMo-Plus}, a benchmark
for assessing cognitive memory under cue--trigger semantic disconnect,
where models must retain and apply latent constraints across long
conversational contexts.
We further show that conventional string-matching metrics and explicit
task-type prompting are misaligned with such scenarios, and propose a
unified evaluation framework based on constraint consistency.
Experiments across diverse backbone models, retrieval-based methods, and
memory systems demonstrate that cognitive memory remains challenging and
reveals failures not captured by existing benchmarks.
Our code and evaluation framework are publicly available at:
\url{https://github.com/xjtuleeyf/Locomo-Plus}.
\end{abstract}

\begin{figure}[t]
  \includegraphics[width=\columnwidth]{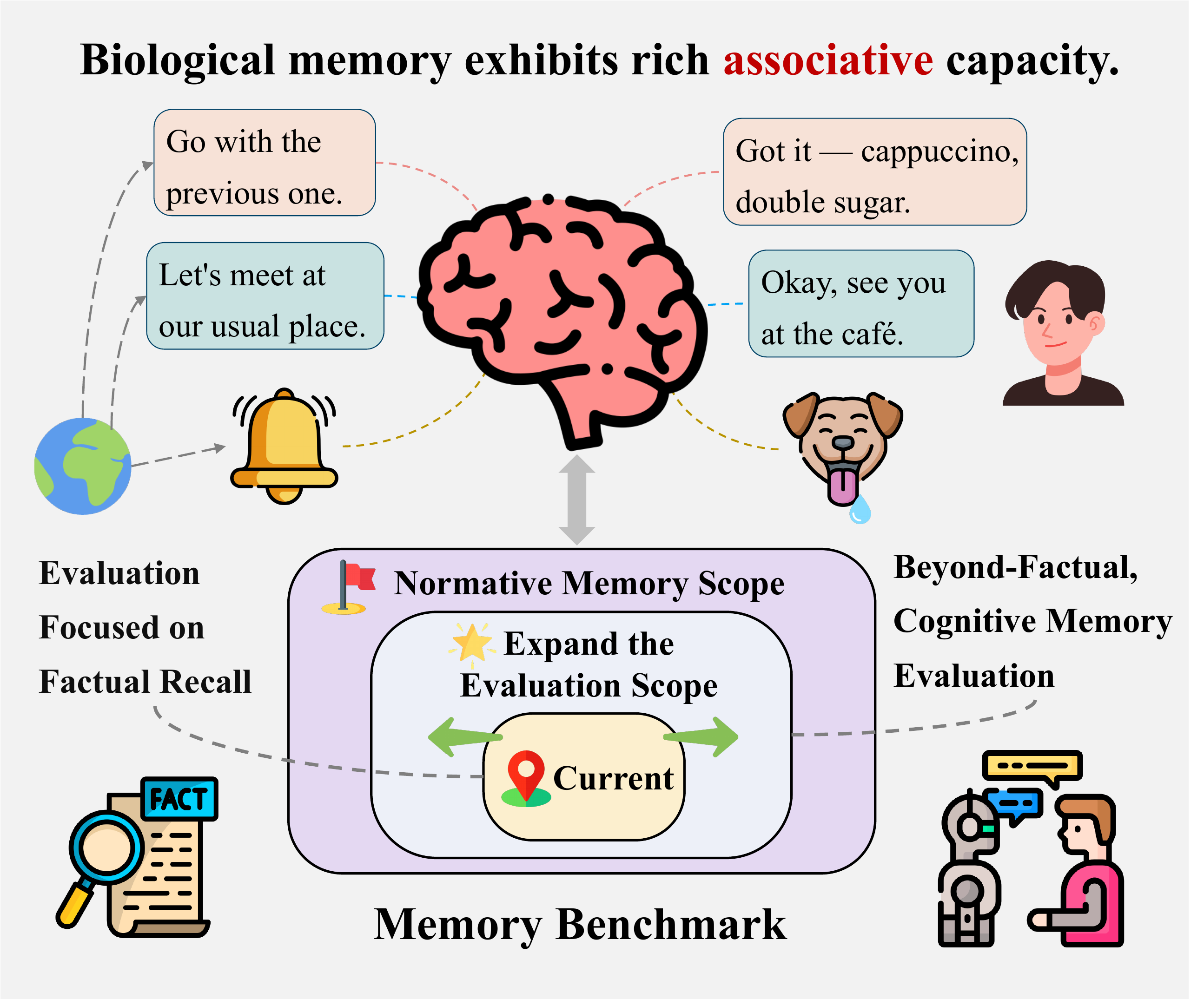}
  \caption{Illustration of the gap between factual memory evaluation and
the richer associative nature of biological memory, motivating the
expansion toward beyond-factual cognitive memory benchmarks.}
  \label{fig:overview}
\end{figure}

\section{Introduction}

Long-term conversational memory is becoming a core capability of modern chatbot systems.
As large language models are increasingly deployed as persistent assistants rather than single-turn responders,
they are expected to remember prior interactions, adapt to user context, and behave consistently over time.
This trend is reflected in both industrial agent designs and recent academic surveys on memory-augmented LLMs
\cite{openai_memory_controls_2025,claude_memory_blog_2025,wu2025human}.
Effective memory reduces repeated cold starts, supports personalization, and is essential for practical conversational AI.\cite{zhang2025survey,hua2025context}

Existing benchmarks have made progress toward evaluating long-term memory by extending question answering to long dialogue contexts
\cite{reddy2019coqa,xu2022beyond,maharana2024evaluating,pan2025memory}.
In particular, LoCoMo evaluates memory under long-context interference and diverse reasoning demands, including multi-hop, temporal, commonsense, and adversarial queries
\cite{maharana2024evaluating}.
These benchmarks are valuable for revealing performance degradation as dialogue length increases.
However, as illustrated in Figure~\ref{fig:overview}, they largely equate conversational memory with explicit factual recall,
where relevant information is clearly stated earlier and later queried with strong semantic alignment.
This formulation enables scalable evaluation, but captures only a limited subset of memory behaviors in natural conversations.

In realistic interactions, memory often constrains behavior rather than answering explicit factual queries.\cite{zheng2025flexible}
Information is frequently conveyed implicitly, and later responses depend on inferred goals or constraints,
a view consistent with findings from cognitive science and neuroscience
\cite{amodio2019social,hoskins2024ai}.
Consider a simple example: earlier in a conversation, a user states that they are preparing for an important exam and want to minimize distractions.
Much later, after many unrelated turns, the user asks, ``Should I start watching that new TV series everyone is talking about?''
There is no explicit fact to recall, and multiple answers may appear reasonable in isolation.
Yet appropriate behavior depends on whether the assistant remembers the earlier goal and recognizes the conflict with the new request.
Such cases cannot be evaluated by surface-level matching, but require assessing consistency with implicit conversational constraints.
Existing benchmarks do not meaningfully test this setting.

In this work, we address this gap by revisiting both benchmark design and evaluation.
We introduce \textsc{LoCoMo-Plus}, a benchmark that targets beyond-factual conversational memory by constructing long-context instances
where correct behavior depends on retaining and applying implicit constraints under cue--trigger semantic disconnect.
We further show that common evaluation practices---such as task-disclosed prompting and string-matching metrics---can be systematically misleading,
even for factual memory, by conflating memory fidelity with prompt adaptation and generation style
\cite{novikova2017we,post2018call}.
To address this issue, we propose a unified evaluation framework based on constraint consistency rather than surface overlap.
Our experiments demonstrate that current LLMs and memory-augmented systems struggle severely under this setting,
indicating that meaningful progress on conversational memory requires rethinking benchmarks and evaluation.

In summary, our contributions are as follows:
\begin{itemize}
    \item We identify a critical limitation in existing conversational
    memory benchmarks, which primarily focus on explicit factual recall
    and fail to capture beyond-factual cognitive memory grounded in
    implicit constraints.
    \item We introduce \textbf{LoCoMo-Plus}, a benchmark designed to
    evaluate cognitive memory under \emph{cue--trigger semantic
    disconnect}, where models must retain and apply latent constraints
    across long conversational contexts.
    \item We revisit conversational memory evaluation and propose a
    unified, constraint-consistency-based framework, supported by
    extensive experiments across diverse backbone models,
    retrieval-based methods, and memory systems.
\end{itemize}

\section{Related Work}

\subsection{Benchmarks for Long-Term Memory}

Existing benchmarks and methods for long-term conversational memory mainly
encourage models to retain and retrieve information that is explicitly
stated in earlier dialogue turns.
Conversational QA and extended dialogue benchmarks typically rely on
semantic alignment between past context and the current query, framing
memory as a retrieval problem
\cite{reddy2019coqa,xu2022beyond,maharana2024evaluating,pan2025memory}.
Similarly, memory-augmented conversational agents often extract,
summarize, or retrieve salient content from dialogue history to generate
history-aware responses.~\cite{packer2023memgpt,zhong2024memorybank,rasmussen2025zep,llamaindex_chat_memory_buffer_2023,langchain_conversation_summary_memory_2023}
Under this common setup, relevant information is assumed to be directly
referencable or semantically matchable, which limits coverage of more
implicit conversational signals, such as user state, goals, or
preferences.

\subsection{Evaluation Metrics and Comparability}
Conversational memory is commonly evaluated using generation-based metrics
such as BLEU, ROUGE, exact match, and token-level F1
\cite{papineni2002bleu,lin2004rouge,post2018call,rajpurkar2018know}.
However, these metrics are known to poorly reflect semantic correctness
and behavioral validity in open-ended generation
\cite{novikova2017we,post2018call,bulian2022tomayto}.
This issue becomes more severe in long-term conversational settings,
where multiple responses can satisfy the same memory requirement while
differing in surface form.
Recent work explores LLM-based judges to improve comparability across
models and generation styles
\cite{zheng2023judging,li2024llms,li2025generation}, but evaluation still
largely focuses on explicit answer correctness.
Consequently, the assessment of implicit and cognitively grounded memory
behaviors remains limited.




\begin{figure*}[t]
  \includegraphics[width=\textwidth]{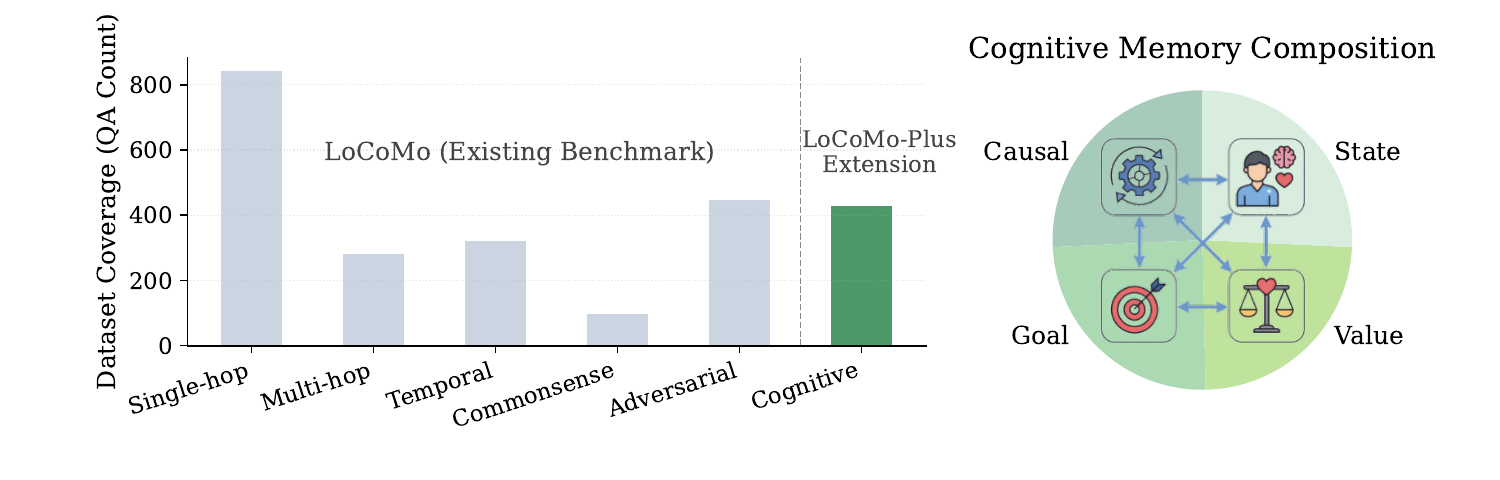}
  \caption{
Cognitive memory in LoCoMo-Plus.
Left: distribution of original LoCoMo question types and the additional
cognitive memory QA instances introduced in LoCoMo-Plus.
Right: cognitive memory is decomposed into four latent constraints—causal,
state, goal, and value.
}
  \label{fig:Cognitive_Memory}
\end{figure*}
\vspace{-0.8em}

\section{Problem Definition}
\label{sec:problem_definition}

We study the evaluation of long-term conversational memory in LLM-based
systems.
In realistic interactions, an agent must not only recall previously
mentioned information, but also retain and utilize knowledge accumulated
over extended dialogue histories to guide future behavior.
This challenge is amplified in long-context settings, where relevant
signals are sparse, temporally distant, and interleaved with irrelevant
content.

Formally, given a multi-turn interaction history
$\mathcal{H} = \{u_1, a_1, \dots, u_t, a_t\}$ and a subsequent user query
$q_{t+1}$, a system produces a response $a_{t+1}$.
The evaluation objective is to assess whether the system effectively
leverages information from $\mathcal{H}$ when generating $a_{t+1}$.

\paragraph{Level-1 Factual Memory.}
Most existing benchmarks evaluate \textbf{Level-1 Factual Memory}, where
relevant information is explicitly stated in the interaction history and
can be directly recalled or reasoned over.
This category includes both localized object-centric facts and
event-oriented episodic information.
Each query admits a well-defined ground-truth answer $y^\ast$, and
correctness is typically determined by string-based or semantic
similarity between $a_{t+1}$ and $y^\ast$.

\paragraph{Level-2 Cognitive Memory.}
In contrast, realistic conversations often depend on \emph{implicit
constraints} inferred from prior interactions, such as user state,
goals, preferences, values, or causal context.
We refer to the ability to retain and apply these latent,
behaviorally constraining signals as \textbf{Level-2 Cognitive Memory}.

Unlike Level-1 memory, Level-2 memory does not admit a single ground-truth
response.
Instead, the interaction history induces a latent constraint $c$ that
restricts the space of behaviorally valid outputs:
\[
\mathcal{A}_c = \{ a \mid a \text{ is consistent with } c \}.
\]
A response is considered correct if it lies within $\mathcal{A}_c$,
regardless of surface form.

As illustrated in Figure~\ref{fig:Cognitive_Memory}, we further decompose
cognitive memory into four interacting components: \emph{causal},
\emph{state}, \emph{goal}, and \emph{value}, which jointly shape
appropriate behavior in long-term conversations.

\paragraph{Dataset Distribution and Coverage.}
We construct a benchmark with explicit coverage over both factual and
cognitive memory behaviors.
As shown in Figure~\ref{fig:Cognitive_Memory}, cognitive cases in
LoCoMo-Plus are intentionally limited and treated as a higher-level
evaluation axis rather than fine-grained factual variants.
Due to their higher generation and verification cost, we prioritize
diagnostic coverage over scale.
Representative cognitive memory cases are provided in
Appendix~\ref{app:cognitive-examples}.

\begin{figure*}[t]
  \includegraphics[width=\textwidth]{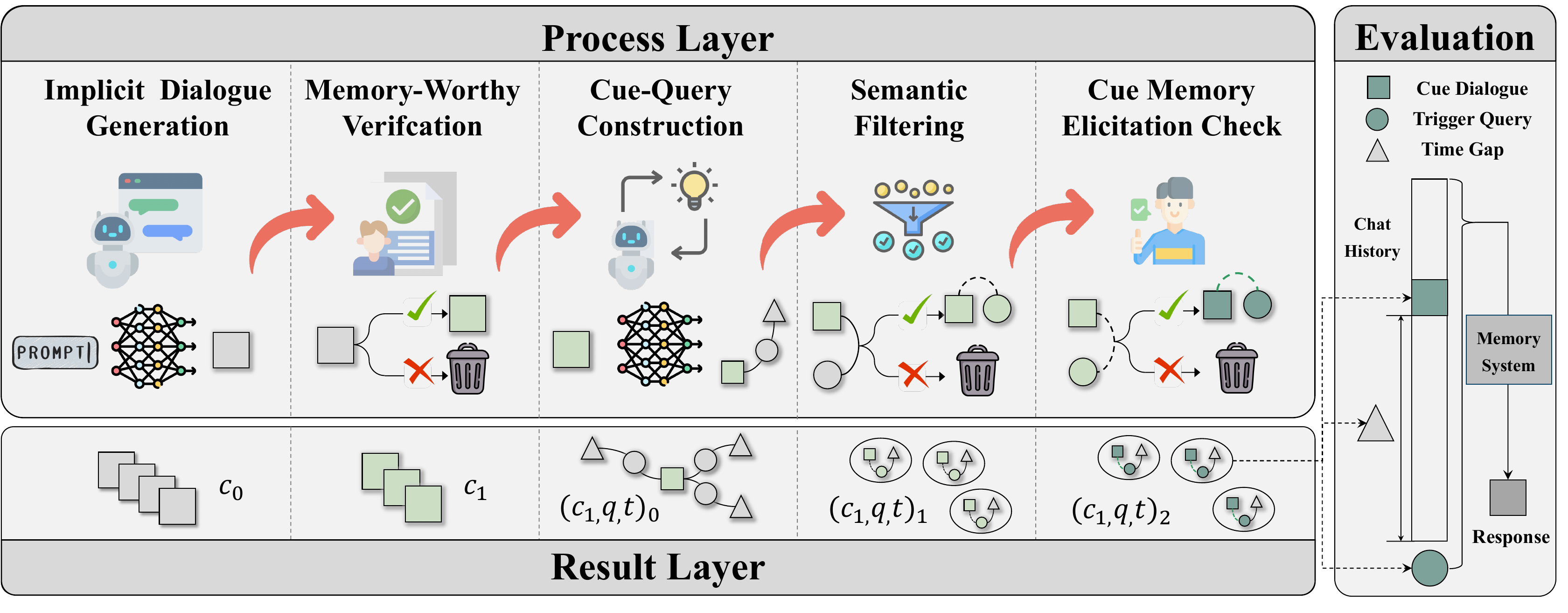}
  \caption{LoCoMo-Plus benchmark construction pipeline with aligned
process and result layers.}
  \label{fig:pipeline}
\end{figure*}

\section{Benchmark Construction}
\label{sec:construction}

LoCoMo-Plus is designed to evaluate \emph{cognitive memory} under
\emph{cue--trigger semantic disconnect}, where a system must preserve and
apply latent user- or speaker-specific information (e.g., inferred
state, goal, or value) even when the downstream query is not
semantically similar to the original cue.
Unlike original LoCoMo instances that primarily test explicit factual
recall, LoCoMo-Plus instances are constructed from scratch as
\emph{cue--trigger pairs} and then embedded into long conversational
trajectories, preserving realistic dialogue structure and long-context
interference.
Figure~\ref{fig:pipeline} illustrates the overall construction pipeline,
which incrementally transforms raw cue dialogues into validated
cue--trigger instances through a sequence of generation, filtering, and
validation stages.

\subsection{Implicit Cue Dialogue Generation}

We first prompt an LLM to generate short dialogue snippets that
implicitly convey memory-relevant information about one participant.
The conveyed information reflects underlying state, goal, preference,
constraint, or value, and is expressed in natural conversational form
rather than as explicit facts.
This stage produces a pool of candidate cue dialogues $c_0$.

\subsection{Memory-Worthy Verification}

Generated cues are manually verified to retain only
\emph{memory-worthy} instances.
A cue is considered memory-worthy if it conveys persistent or
behaviorally constraining information, cannot be trivially inferred from
local context alone, and would plausibly benefit a conversational
assistant if remembered.
Cues failing these criteria are discarded, yielding a verified set
$c_1$.

\subsection{Cue--Trigger Query Construction}

Given a verified cue $c_1$, we prompt an LLM to generate a downstream
\emph{trigger query} $q$ whose correct resolution depends on the cue,
while maintaining low surface-level semantic similarity.
The query is underspecified in isolation, such that multiple responses
may appear reasonable without the cue, but only cue-consistent responses
are valid.
Each cue--query pair is additionally assigned a temporal gap indicator
$t$, specifying the distance between cue and trigger when embedded into a
long dialogue.
This produces preliminary tuples $(c_1, q, t)_0$.

\subsection{Semantic Filtering}

To avoid shortcut solutions based on surface-level overlap,
we filter out cue–query pairs with high lexical or semantic similarity.
Specifically, we apply BM25~\cite{Amati2009} and MPNet-based~\cite{song2020mpnet} similarity scoring to remove
cases where the cue is repeated, paraphrased, or directly recoverable
from the query alone.
The remaining pairs form the filtered set $(c_1, q, t)_1$.

\subsection{Cue Memory Elicitation Validation}

A final round of human validation ensures that each remaining
cue--trigger pair genuinely elicits memory usage.
Annotators verify that producing a helpful response requires recalling
and applying information from the cue through an implicit constraint,
rather than relying on surface similarity.
Only validated instances are retained, forming the final set
$(c_1, q, t)_2$.

\subsection{Insertion into LoCoMo Dialogues}

Each validated instance $(c_1, q, t)_2$ is embedded into a selected long
dialogue trajectory from LoCoMo by inserting the cue and placing the
corresponding trigger query after a gap consistent with $t$.
The resulting examples require models to retain and utilize cue
information despite intervening turns and distractors.

\paragraph{Implementation Details.}
Additional details on models, prompting strategies, annotation
guidelines, and filtering criteria are provided in Appendix~\ref{app:benchmark_details}.

\section{Evaluation Framework}
\label{sec:evaluation}

\begin{figure}[t]
  \includegraphics[width=\columnwidth]{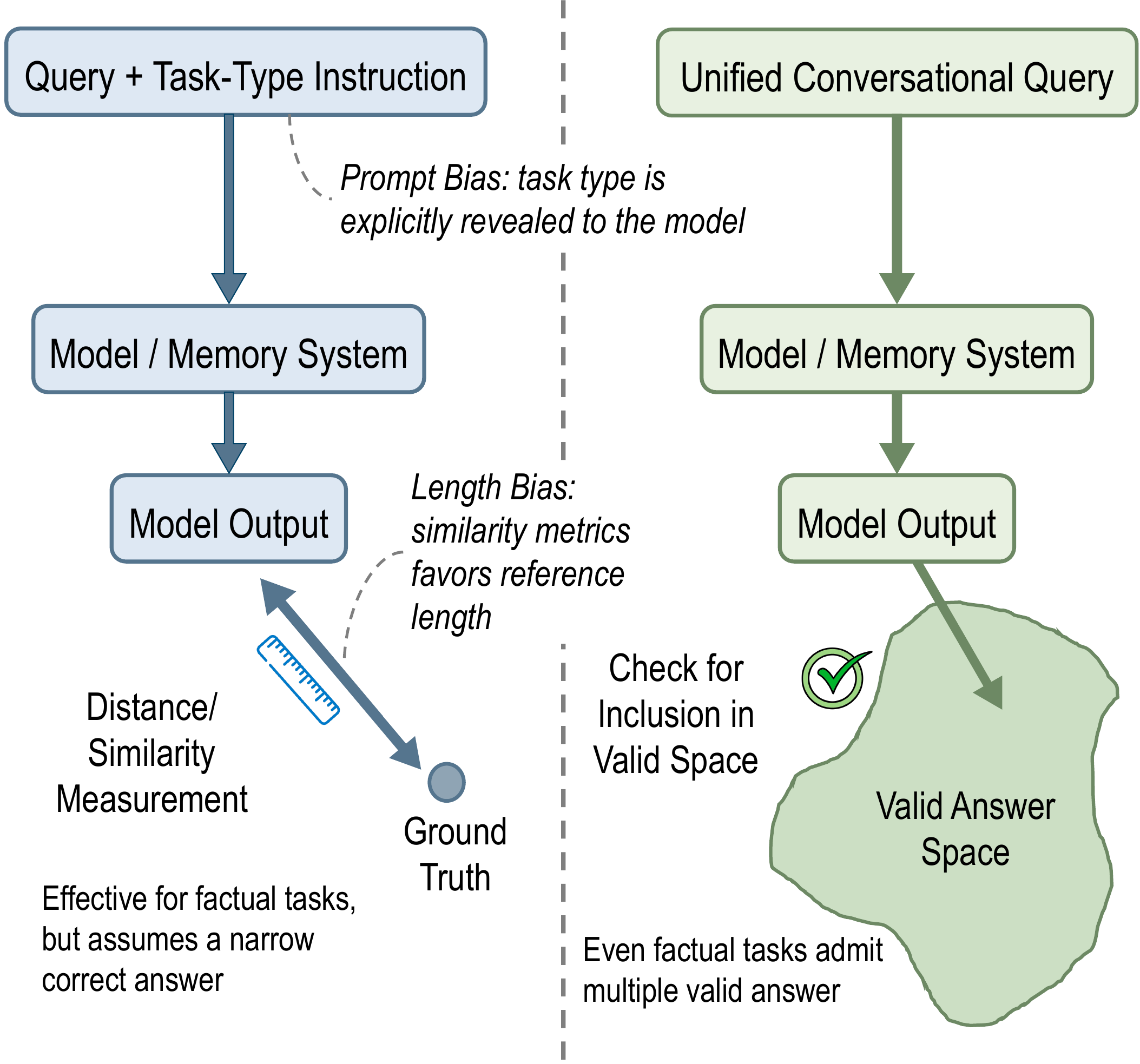}
  \caption{Conceptual comparison between the existing evaluation framework
and the proposed evaluation paradigm for conversational memory.}
  \label{fig:evaluation_paradigm}
\end{figure}

Existing LoCoMo-style benchmarks evaluate long-term conversational
memory by extending question answering paradigms to long dialogue
contexts.~\cite{maharana2024evaluating}
In practice, queries are augmented with explicit task-type instructions
(e.g., indicating that a query tests memory), and model outputs are
evaluated using generation-based metrics such as exact match,
token-level F1, or n-gram overlap with a reference answer.~\cite{rajpurkar2018know,lin2004rouge,post2018call}
This design enables scalable evaluation under long context, but also
introduces strong inductive biases.~\cite{perez2021true,min2022rethinking}

When applied to modern large language models, this evaluation setup
systematically distorts what is measured as memory performance.
By disclosing task intent at the input side and relying on surface-form
matching at the output side, the evaluation increasingly reflects
models’ adaptation to task prompts and generation styles rather than
their ability to retain and apply conversational context.~\cite{perez2021true,min2022rethinking,shi2023large}
As a result, the reported performance can be misleading, undermining
both cross-task and cross-model comparability.
Figure~\ref{fig:evaluation_paradigm} illustrates these effects.

\subsection{Input-Side Assumptions: Task Disclosure}

On the input side, LoCoMo-style evaluation explicitly specifies the task
type before each query.
This formulation conditions model behavior on task identity, encouraging
task-specific response strategies instead of implicit recall from prior
dialogue.~\cite{perez2021true,min2022rethinking}
Consequently, different memory tasks elicit qualitatively different
answering behaviors even for the same model.

This design breaks comparability across task categories.
Performance differences may reflect sensitivity to task prompts rather
than differences in memory capability.
Moreover, such task disclosure does not reflect natural conversational
use, where users do not announce that a query requires recalling earlier
information.~\cite{reddy2019coqa,xu2022beyond}

\subsection{Output-Side Assumptions: Generation-Based Metrics}

On the output side, LoCoMo relies on string-matching-based metrics to
compare model outputs with a reference answer.
These metrics assume that correctness can be determined by surface-level
overlap.~\cite{bulian2022tomayto,novikova2017we}

In conversational memory settings, this assumption does not hold.
Multiple valid responses may satisfy the same memory constraint while
receiving different scores.~\cite{bulian2022tomayto,post2018call}
At the same time, modern LLMs exhibit diverse generation behaviors in
terms of verbosity, reasoning style, and response length.
As a result, generation-based metrics systematically bias comparisons
across models, conflating memory fidelity with surface realization.
~\cite{zheng2023judging,li2024llms}

\subsection{Constraint-Consistency Evaluation}

To address these limitations, we reformulate LoCoMo-style evaluation
around \emph{constraint consistency}.
Queries are presented as natural continuations of the dialogue without
explicit task disclosure.
Rather than scoring surface-form overlap against a single reference,
we evaluate whether a model’s response satisfies the implicit constraint
induced by the cue.

\begin{table*}[t]
\centering
\small
\setlength{\tabcolsep}{5pt}
\begin{tabular}{lcccccccc}
\toprule
\textbf{Method}
& \multicolumn{6}{c}{\textbf{LoCoMo (Factual Memory)}} 
& \textbf{LoCoMo-Plus}
& \textbf{Gap} \\
\cmidrule(lr){2-7}

& single-hop
& multi-hop
& temporal
& commonsense
& adversarial
& average
& 
& \\

\midrule
\multicolumn{9}{l}{\textit{Open Source LLM}} \\

\midrule
Qwen2.5-3B-Instruct   & 68.25 & 38.65 & 18.38 & 48.44 & 11.69 & 42.20 & 10.82 & 31.38 \\
Qwen2.5-7B-Instruct   & 70.72 & 39.54 & 21.81 & 37.50 & 20.22 & 45.31 & 9.57  & 35.74 \\
Qwen2.5-14B-Instruct  & 76.33 & 48.23 & 38.94 & 57.29 & 68.09 & 63.45 & 19.24 & 44.21 \\
Qwen3-4B              & 69.52 & 46.10 & 33.33 & 55.21 & 48.76 & 54.91 & 15.70 & 39.21 \\
Qwen3-8B              & 69.34 & 43.79 & 39.88 & 59.90 & 53.48 & 56.86 & 17.68 & 39.18 \\
Qwen3-14B             & 65.96 & 46.45 & 53.89 & 59.38 & 60.45 & 59.65 & 19.09 & 40.56 \\

\midrule
\multicolumn{9}{l}{\textit{Close Source LLM}} \\
\midrule
gpt-5-nano          & 75.00 & 54.08 & 50.16 & 73.96 & 17.53 & 54.96 & 14.84 & 40.12 \\
gpt-4.1             & 80.30 & 53.90 & 58.88 & 72.92 & 37.30 & 62.21 & 18.63 & 43.58 \\
gpt-4o              & 78.13 & 52.30 & 45.79 & 69.79 & 48.99 & 62.99 & 21.05 & 41.94 \\
gemini-2.5-flash    & 77.71 & 54.26 & 66.04 & 66.67 & 65.84 & 69.25 & 24.67 & 44.58 \\
gemini-2.5-pro      & 77.83 & 52.48 & 73.83 & 63.54 & 73.03 & 71.78 & 26.06 & 45.72 \\

\midrule
\multicolumn{9}{l}{\textit{RAG-based Methods (GPT-4o)}} \\
\midrule
Text-ada-embedding-002  & 40.00 & 16.73 & 37.81 & 15.73 & 49.44 & 37.38 & 13.91 & 23.47 \\
Text-embedding-small    & 39.17 & 17.79 & 34.69 & 14.61 & 51.90 & 37.23 & 12.29 & 24.94 \\
Text-embedding-large    & 49.76 & 22.78 & 40.00 & 21.35 & 59.73 & 45.32 & 15.55 & 29.77 \\

\midrule
\multicolumn{9}{l}{\textit{Memory Systems (GPT-4o)}} \\
\midrule
Mem0            
& 80.20 
& 48.10 
& 39.40 
& 66.20 
& 30.50 
& 57.24 
& 15.80 
& 41.44 \\
SeCom           
& 77.60 
& 50.90 
& 42.30 
& 71.40 
& 31.80 
& 57.53 
& 14.90 
& 42.63 \\
A-Mem           
& 76.90 
& 55.60 
& 49.30 
& 68.10 
& 35.20 
& 59.64 
& 17.20 
& 42.44 \\

\bottomrule
\end{tabular}
\caption{Overall performance of a wide range of models and memory
systems on LoCoMo (factual memory) and LoCoMo-Plus (cognitive memory),
reported across task categories. The Gap column indicates the
performance drop from LoCoMo to LoCoMo-Plus.}
\label{tab:overall}
\end{table*}

Under this formulation, correctness is defined as membership in a valid
response space, allowing multiple acceptable realizations.
By jointly removing task disclosure and decoupling evaluation from
generation style~\cite{zheng2023judging,li2025generation},
we further decouple task formulation from judgment criteria and employ
task-specific evaluation aligned with the reasoning demands of each task.
This unified-input, differentiated-judgment paradigm enables coherent
and comparable assessment across memory tasks and model families,
covering both factual and cognitive memory.

Empirical results supporting these claims are presented in section~\ref{bias_analysis}.

\section{Experiments}

\subsection{Experimental Setup and Model Coverage}
\label{sec:exp_setup}

We evaluate a broad range of conversational memory methods on both
\textbf{LoCoMo} and \textbf{LoCoMo-Plus}, corresponding to the
factual memory and cognitive memory
regimes defined in Section~\ref{sec:problem_definition}.
All methods are evaluated under identical dialogue contexts and query
formats, and are assessed using the same output-side evaluation
protocol.

\paragraph{Method Coverage.}
We evaluate methods spanning four categories of conversational memory
approaches.

\textbf{Open-Source LLMs.}
We evaluate instruction-tuned open-source language models that rely solely on
native context modeling, where the complete conversation history is provided
as input without any external retrieval or memory mechanism~\cite{team2024qwen2,yang2025qwen3}.

\textbf{Closed-Source LLMs.}
We evaluate proprietary large language models under the same full-context
input and output protocols, serving as strong reference baselines for
context-only conversational modeling~\cite{achiam2023gpt,team2023gemini}.

\textbf{RAG-based Methods.}
Retrieval-augmented baselines retrieve a fixed set of the top-5 most relevant
dialogue segments from an external embedding-based memory store conditioned on
the current query, and append them to the prompt for response generation.
We evaluate retrieval using three OpenAI embedding models.~\cite{openai_embeddings}.

\textbf{Memory Systems.}
We evaluate three state-of-the-art conversational memory systems:
\emph{A-Mem}~\cite{xu2025mem}, which maintains and retrieves structured
long-term memories through adaptive memory construction and retrieval;
\emph{Mem0}~\cite{chhikara2025mem0}, which provides a production-oriented
memory abstraction with scalable long-term storage and retrieval; and
\emph{SeCom}~\cite{pan2025memory}, which constructs segment-level memory units
with compression-based denoising to improve retrieval accuracy in long-term
conversations.

\subsection{Overall Performance and Main Findings}

Table~\ref{tab:overall} summarizes the overall performance of a broad
range of conversational memory methods on both LoCoMo and LoCoMo-Plus.
Across backbone models, retrieval-augmented pipelines, and dedicated
memory systems, a single and consistent pattern emerges:
\textbf{LoCoMo-Plus remains challenging for all evaluated methods}.

\paragraph{LoCoMo-Plus exposes a persistent and unresolved challenge.}
Regardless of backbone strength, architectural design, or memory
mechanism, all methods exhibit a substantial performance gap between
LoCoMo and LoCoMo-Plus.
This gap persists even for the strongest and most recent LLMs, as well
as for systems explicitly designed to enhance long-term memory.
The consistency of this degradation suggests that the difficulty of
LoCoMo-Plus arises from the task formulation itself—namely, preserving
and applying implicit constraints under cue--trigger semantic
disconnect—rather than from specific modeling or engineering choices.
Taken together, these results indicate that cognitive conversational
memory, as instantiated in LoCoMo-Plus, remains an open problem for
current approaches.

\paragraph{Additional observations.}
While methods show noticeable performance variation on the original
LoCoMo benchmark, these variations become markedly less pronounced under
the LoCoMo-Plus setting.
Across diverse modeling choices, relative performance differences are
compressed, reflecting a convergence toward uniformly low performance
when implicit constraint preservation is required.

\subsection{Evaluation Bias Analysis}
\label{bias_analysis}

\begin{figure}[t]
  \includegraphics[width=\columnwidth]{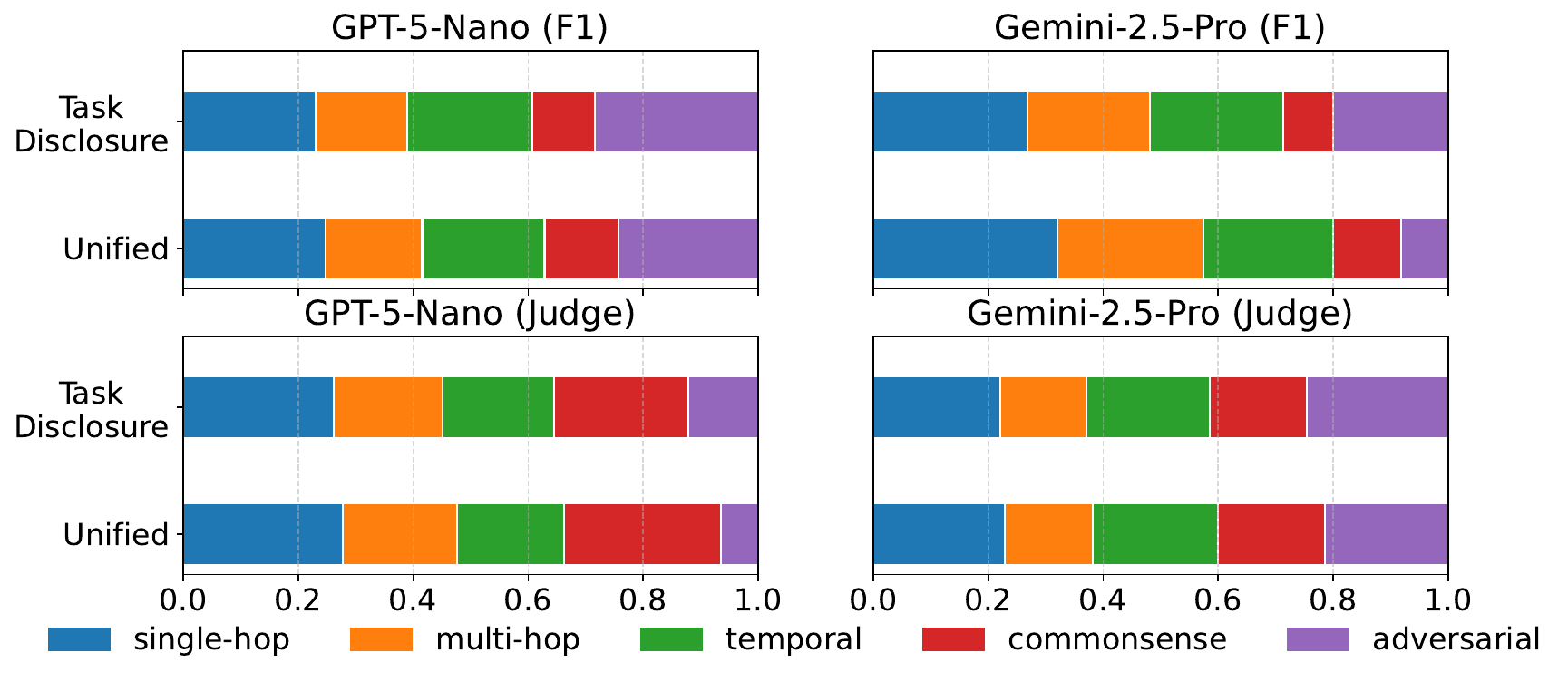}
  \caption{Comparison of task-disclosed and unified dialogue inputs
across task types, evaluated with different output-side
assessment methods and model families.}
  \label{fig:prompt_bias}
\end{figure}

\begin{figure}[t]
  \includegraphics[width=\columnwidth]{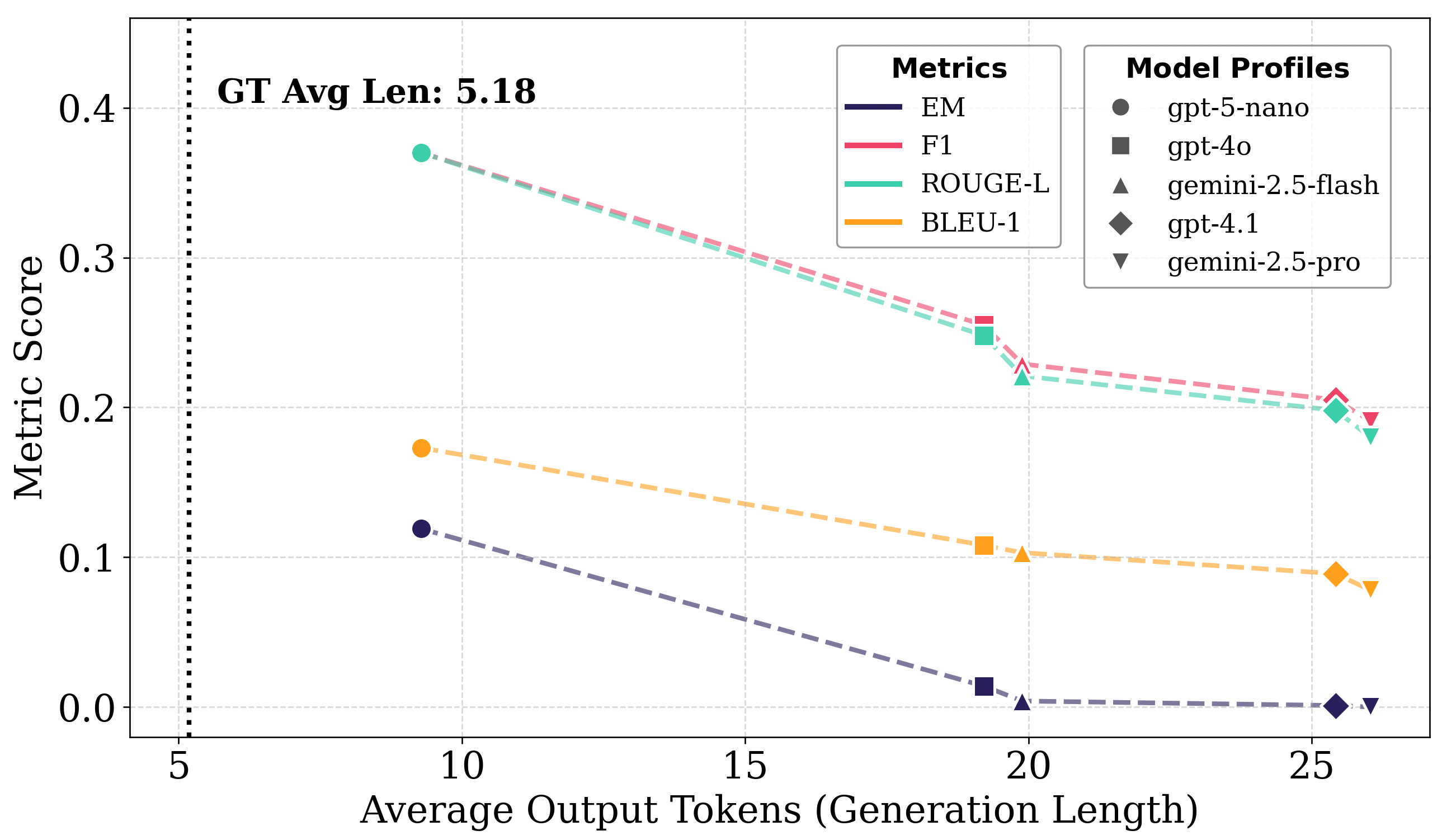}
  \caption{Traditional metric scores under different average generation lengths.
Each line corresponds to a metric, with markers indicating different models.
The dashed vertical line denotes the average ground-truth length (5.18 tokens).}
  \label{fig:length_bias}
\end{figure}

We empirically verify that the evaluation assumptions discussed in
Section~\ref{sec:evaluation} lead to systematic biases when applied to
modern large language models.
Figure~\ref{fig:prompt_bias} and Figure~\ref{fig:length_bias} illustrate
how prompt disclosure and generation-based metrics distort measured
memory performance in practice.

\paragraph{Prompt Bias.}
To isolate the effect of task disclosure, we fix the base model and evaluation
metric and compare task-wise ability distributions under task-disclosed and
unified conversational queries (Figure~\ref{fig:prompt_bias}).
Across both models and evaluation standards, explicitly revealing task
identity leads to a pronounced shift in the task-wise performance
distribution.
Notably, the distortion concentrates on temporal reasoning and
adversarial (hallucination-related) tasks, which receive disproportionately
higher scores under task-disclosed evaluation.
These task categories are central to LoCoMo-style benchmarks, suggesting
that reported gains may partly reflect sensitivity to task prompts rather
than stable memory behavior.
This analysis indicates that task disclosure can bias the measured
ability profile, making task-wise comparisons under LoCoMo-style evaluation
less reliable.

\paragraph{Length Bias.}
We analyze multiple traditional generation-based metrics across different
closed-source models and observe a clear dependence between metric scores
and average output length.
As shown in Figure~\ref{fig:length_bias}, EM, F1, BLEU~\cite{papineni2002bleu}, and ROUGE~\cite{lin2004rouge} all vary
systematically with the number of generated tokens, with scores peaking
near the average ground-truth length and degrading as outputs become
shorter or longer.
This behavior is expected given the formulation of these metrics, which
reward surface-level overlap and implicitly favor outputs whose length
closely matches the reference.
As a result, models with different generation styles are penalized or
favored based on length alone, regardless of semantic correctness,
introducing a systematic length bias in cross-model comparison.


\begin{table}[t]
\centering
\small
\setlength{\tabcolsep}{6pt}
\begin{tabular}{lccc}
\toprule
 & Human$_1$ & Human$_2$ & LLM Judge \\
\midrule
Human$_1$   & --    & 0.903 & 0.801 \\
Human$_2$   & 0.903 & --    & 0.820 \\
LLM Judge   & 0.801 & 0.820 & --    \\
\bottomrule
\end{tabular}
\caption{Pairwise agreement between two human annotators and the LLM judge
(\texttt{gemini-2.5-flash}). Agreement is computed using normalized
pairwise agreement scores under a shared evaluation protocol.}
\label{tab:judge_agreement}
\end{table}

\begin{table}[t]
\centering
\small
\setlength{\tabcolsep}{6pt}
\begin{tabular}{lcccccc}
\toprule
 & \multicolumn{3}{c}{\textbf{Qwen2.5}} & \multicolumn{3}{c}{\textbf{Qwen3}} \\
\cmidrule(lr){2-4} \cmidrule(lr){5-7}
\textbf{Judge} 
& \textbf{3B} & \textbf{7B} & \textbf{14B}
& \textbf{4B} & \textbf{8B} & \textbf{14B} \\
\midrule
Judge 1 & 42.20 & 45.31 & 63.45 & 54.91 & 56.89 & 59.65 \\
Judge 2 & 45.24 & 48.64 & 62.77 & 56.94 & 58.54 & 60.37 \\
\midrule
$|\Delta|$ & 3.04 & 3.33 & 0.68 & 2.03 & 1.65 & 0.72 \\
\bottomrule
\end{tabular}
\caption{Score stability across different judge backbones on the same set of response models.
$|\Delta|$ denotes the absolute score difference between Judge~1 and Judge~2.}
\label{tab:judge_stability}
\end{table}

\begin{figure}[t]
  \includegraphics[width=\columnwidth]{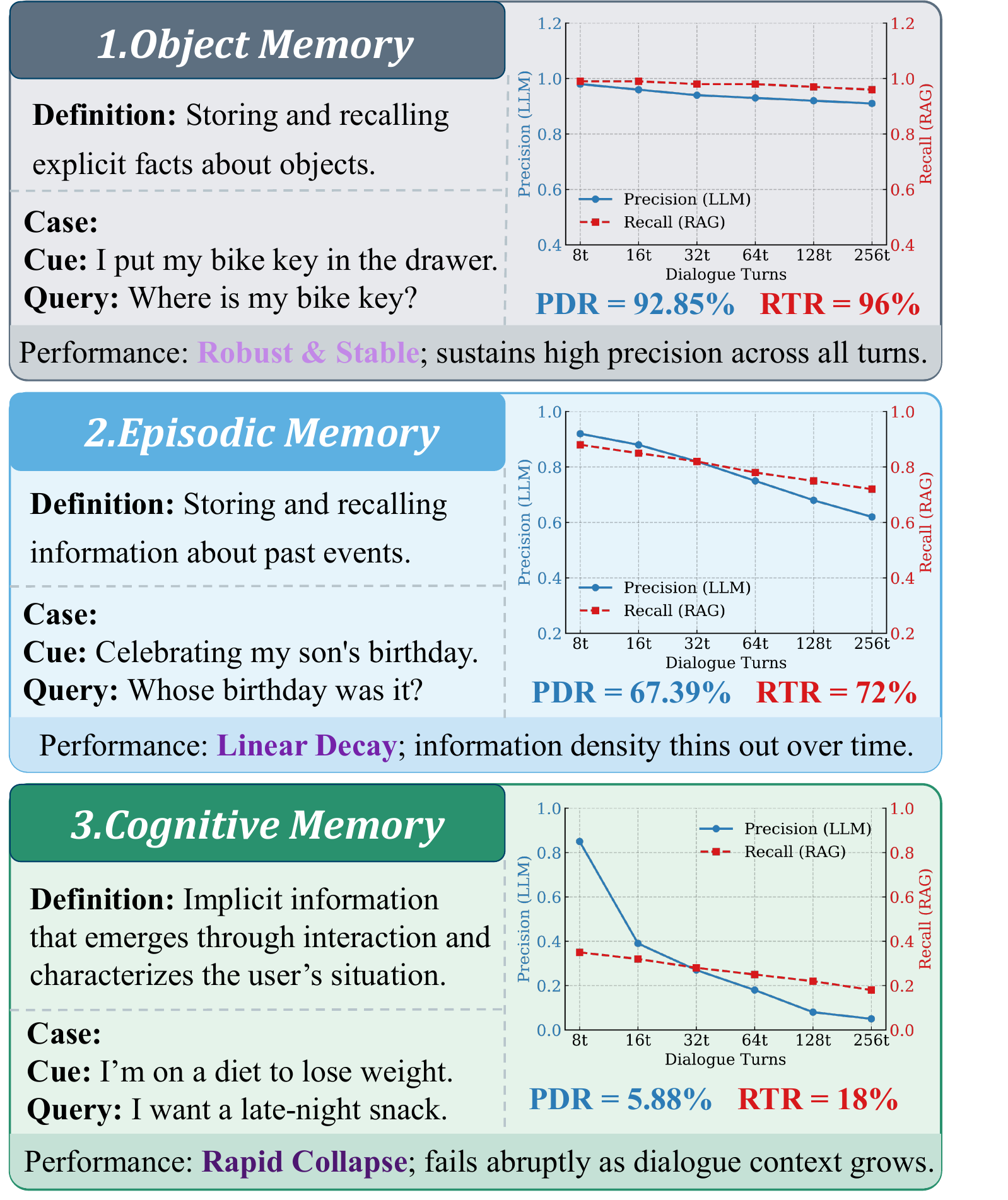}
  \caption{Length sensitivity across (object / episodic / cognitive) memory under increasing dialogue context.}
  \label{fig:experiments}
\end{figure}

\subsection{Reliability of LLM-as-a-Judge.}

Notably, judging whether a response satisfies a given conversational
constraint is substantially easier than generating the response itself.
Prior studies have shown that even moderately strong general-purpose LLMs
can perform reliably on such judgment tasks, with high consistency against
human annotations.~\cite{zheng2023judging,thakur2025judging}
Nevertheless, we explicitly quantify judge robustness in our setting.
We evaluate the reliability of LLM-based judging along two dimensions:
alignment with human judgment and stability across judge backbones.

Table~\ref{tab:judge_agreement} shows strong agreement between the two human annotators, as well as high agreement between each annotator and the LLM judge, indicating that
the LLM judge closely aligns with human evaluation.

Table~\ref{tab:judge_stability} compares scores produced by two different
judge backbones (\texttt{Gemini-2.5-Flash} and \texttt{GPT-4o}) on the same
set of model responses.
Across multiple response models, score differences remain small, suggesting
that evaluation outcomes are stable with respect to the choice of judge.

Details of the judge prompt, source annotations, and agreement statistics
are provided in Appendix~\ref{app:judge_details}.

\subsection{Length Sensitivity Analysis}

Figure~\ref{fig:experiments} analyzes length sensitivity across object,
episodic, and cognitive memory by varying the number of dialogue turns.
For each memory type, we evaluate 100 representative cases following the
distinction in Section~\ref{sec:problem_definition}.

Object memory remains robust as context grows, reflecting the stability of
highly localized factual recall.
Episodic memory exhibits a steady degradation pattern with increasing
dialogue length, indicating increasing difficulty in recovering
temporally distributed factual information.

In contrast, cognitive memory shows rapid performance collapse as context
length increases.
This sharp disparity highlights that Level-2 cognitive memory is
substantially more sensitive to long-context interference than Level-1
factual memory, and is not adequately captured by benchmarks focused on
explicit factual recall.

\section{Conclusion}

We show that long-term conversational memory evaluation fails when reduced to surface-form recall. \textbf{LoCoMo-Plus} reframes memory assessment as constraint consistency, revealing cognitive memory behaviors that existing benchmarks miss and enabling reliable evaluation through evidence-grounded LLM judges.

\section*{Limitations}

This work evaluates beyond-factual conversational memory through
implicit constraints under cue--trigger semantic disconnect.
While LoCoMo-Plus extends the scope of existing memory benchmarks, it is
not intended to cover all forms of human memory or cognitive processes.
The benchmark focuses on conversational settings where latent
constraints influence behavior, but does not model long-term belief
revision, emotional dynamics, or multi-agent memory interactions.
LoCoMo-Plus prioritizes diagnostic value over scale.
Its instances are carefully generated and validated to ensure genuine
memory usage, resulting in a dataset smaller than large-scale factual QA
benchmarks and unsuitable for training or fine-tuning large language
models.
Our evaluation relies on LLM-based judges to assess constraint
consistency.
Although judge reliability is empirically examined, results may remain
sensitive to the choice of judge model and prompt design.
Our experiments are limited to English-language conversations and a
specific set of backbone models, retrieval pipelines, and memory
systems, leaving broader generalization to future work.

\section*{Ethical Considerations}

This work focuses on evaluation and does not introduce new model
training procedures or deploy systems in real-world applications.
LoCoMo-Plus is intended solely for research and diagnostic purposes.
All dialogue content is synthetically generated or adapted from existing
benchmarks and contains no personal, sensitive, or identifiable
information.
The implicit constraints modeled (e.g., goals or preferences) are
abstract and not tied to real users.
While enhanced conversational memory may raise concerns related to
privacy and user control, this work evaluates memory use within a
controlled setting rather than promoting persistent storage of user
data.
Questions regarding how memory should be stored, forgotten, or governed
in deployed systems are beyond the scope of this paper.

\bibliography{custom}

\clearpage

\appendix

\section{Prompting and Generation Details}
\label{app:benchmark_details}

We describe the prompting design and generation parameters used to
construct cue dialogues and trigger queries in our dataset.
These prompts are used exclusively for data generation and do not involve
any evaluation or judgment procedures.

We adopt a structured prompting strategy in which each prompt is
decomposed into semantically annotated components, including role setting,
task logic, constraints, and output format.
This decomposition clarifies the intended function of each prompt segment
and helps ensure consistency and reproducibility across generations.
Cue dialogues are generated as short two-turn conversations that introduce
a salient and recallable memory anchor while remaining naturalistic and
self-contained.

Trigger queries are then generated conditioned on the cue dialogue and
relation type, with explicit constraints encouraging diversity in
perspective, temporal gap, and cognitive angle.
Complete prompt specifications with semantic annotations are provided in
Tables~\ref{tab:cue_prompt} and~\ref{tab:trigger-prompt}.

\paragraph{Generation Parameters.}
Cue dialogues and trigger queries are generated using a set of
commercially available large language models,
including \texttt{gpt-5-nano}, \texttt{gpt-4o}, \texttt{gpt-4.1}~\cite{achiam2023gpt},
\texttt{gemini-2.5-flash}, and \texttt{gemini-2.5-pro}~\cite{team2023gemini}.
All models are used exclusively for data generation in this stage.

To ensure reproducibility across different model APIs, we restrict
generation control to a shared set of supported parameters.
Specifically, we enable stochastic decoding with a non-zero temperature
($T=0.7$) to encourage diversity in linguistic realization and topic
coverage.
We additionally control the maximum generation length (256 tokens) and
the number of generated instances per relation type
(\texttt{num\_samples}=50).

Aside from enforcing the specified structural and output-format
constraints, no additional filtering or post-processing is applied.
All prompting and generation procedures are implemented in our open-source
codebase, which is released alongside this paper to support full
reproducibility.

\section{Judge Design and Evaluation Protocol}
\label{app:judge_details}

For factual and commonsense questions, model responses are often
open-ended and may partially satisfy the information need.
For example, when a question asks what a user ate for dinner and the
reference answer specifies a concrete item (e.g., ``grilled salmon''),
a response such as ``seafood'' captures the correct coarse concept but
misses the required level of specificity.
Treating such responses as fully incorrect would conflate coarse-grained
recall with complete failure.
We therefore retain a three-level label scheme (\texttt{correct},
\texttt{partial}, \texttt{wrong}) to distinguish partial factual recall
from entirely incorrect or hallucinated answers.

In contrast, temporal and adversarial questions admit precise decision
boundaries.
Temporal reasoning requires exact calculation or ordering, where near
misses or approximate answers are not semantically acceptable.
Similarly, adversarial questions evaluate whether the model correctly
refuses to answer or identifies a conflict; any attempt to hallucinate an
answer constitutes a failure.
For these categories, we adopt binary labels to avoid ambiguity and
ensure consistent evaluation.

Finally, cognitive awareness questions assess whether a response
explicitly acknowledges or adapts to a previously stated memory cue.
Because the core criterion is the presence or absence of memory-aware
behavior rather than surface correctness, we likewise use binary labels
to reflect recall success versus omission.
The full task-specific evaluation prompts are summarized in Table~\ref{tab:unified-evaluation-prompts}.

\subsection{Representative Judge Examples}

To further assess the reliability of LLM-based judgment, we analyze cases
where LLM judges and human annotations disagree.
Such disagreements are typically treated as potential failure cases of
LLM-based evaluation and therefore warrant closer inspection.

Table~\ref{tab:representative-metrics-case} presents a representative
example from this analysis.
At first glance, the conflict between human annotation and the LLM
judge’s decision might suggest an error on the part of the judge.
However, a careful re-examination of the dialogue evidence reveals that
the model prediction is fully consistent with the explicit temporal cue
provided in the conversation.
The source of disagreement instead lies in the reference answer, which
encodes an abstract and inconsistent weekday description.

In this case, human annotators rely on the flawed reference and label the
prediction as incorrect.
The LLM judge, by grounding its decision in the dialogue evidence, not
only assigns the correct label but also explicitly identifies the
inconsistency in the reference answer.
This outcome demonstrates that disagreement with human annotation does
not necessarily indicate unreliability of LLM-based judgment.
Rather, such cases can reflect issues in the reference itself, and the
ability of LLM judges to surface and reason about these inconsistencies
provides additional evidence of their reliability.

Complete per-instance human annotations, LLM judge labels, and judge
rationales are released in our code repository to support transparent
inspection and verification.

\begin{table}[t]
\centering
\small
\setlength{\tabcolsep}{6pt}
\renewcommand{\arraystretch}{1.05}
\begin{tabular}{p{0.30\columnwidth} p{0.65\columnwidth}}
\toprule
\textbf{Component} & \textbf{Content (Verbatim)} \\
\midrule

\multicolumn{2}{l}{\textbf{(A) Dialogue Evidence and Response}} \\

\textbf{Question} &
When did Melanie run a charity race? \\

\textbf{Evidence} &
(D2:1) Melanie: Hey Caroline, since we last chatted, I've had a lot of
things happening to me. I ran a charity race for mental health last
\textbf{Saturday} – it was really rewarding. \\

\textbf{Model Prediction} &
Melanie ran a charity race for mental health on \textbf{Saturday}, before May 25,
2023. \\

\midrule
\multicolumn{2}{l}{\textbf{(B) Reference Answer}} \\

\textbf{Ground Truth} &
The \textbf{Sunday} before 25 May 2023 \\

\midrule
\multicolumn{2}{l}{\textbf{(C) Evaluation Outcomes}} \\

\textbf{Human 1 Judge} &
\textbf{Wrong} \\

\textbf{Human 2 Judge} &
\textbf{Wrong} \\

\textbf{LLM Judge} &
\textbf{Correct} \\

\textbf{Judge Reason} &
The prediction correctly identifies the event as occurring on Saturday
based on the dialogue evidence, while the reference answer uses an
inconsistent weekday description. \\

\midrule
\multicolumn{2}{l}{\textbf{(D) Traditional Metrics}} \\

\textbf{Metrics} &
EM = 0.0;\quad Token F1 $\approx$ 0.40;\quad 

BLEU $\approx$ 0.0;\quad
ROUGE-L $\approx$ 0.2 \\
\bottomrule
\end{tabular}
\caption{A representative bias case where the model prediction aligns
with the dialogue evidence but is penalized by human annotation and
surface-form metrics. The LLM-based judge correctly grounds its decision
in the evidence and assigns the correct label.}
\label{tab:representative-metrics-case}
\end{table}

\section{Representative Cognitive Memory Examples}
\label{app:cognitive-examples}

To concretely illustrate how our dataset differs from fact-centric
benchmarks such as LoCoMo, we present several representative cognitive
memory cases sampled from the full dataset.
All examples below are shown verbatim (lightly anonymized) and are drawn
from different cognitive relation types.
Rather than querying explicitly stated facts, these cases require models
to retain and apply implicit personal constraints expressed earlier in
the dialogue.

\paragraph{Example 1: Causal Memory (Habit Change Triggered by an Event).}

\textbf{Cue Dialogue (Causal):}
\begin{quote}
\textit{A: Since my cousin got diagnosed with Type 2 diabetes, I cut
sugary drinks completely out of my diet.} \\
\textit{B: That family scare clearly changed your habits.}
\end{quote}

\textbf{Trigger Query (Months Later):}
\begin{quote}
\textit{A: It’s strange, I barely recognize the person who used to grab
whatever sounded good without thinking about long-term consequences.}
\end{quote}

\textbf{Analysis.}
The trigger does not explicitly mention diabetes or dietary choices.
Correct interpretation requires recalling the earlier causal event and
connecting it to a later shift in self-perception.
This goes beyond factual recall (e.g., ``Why did you stop drinking sugary
drinks?'') and instead evaluates whether the model can implicitly link a
past cause to a present reflection.

\paragraph{Example 2: State-Based Memory (Emotional State Driving Later Behavior).}

\textbf{Cue Dialogue (State):}
\begin{quote}
\textit{A: I got yelled at by my boss for a mistake I can’t stop thinking
about, and I feel hollow.} \\
\textit{B: That’ll pass; give yourself credit for fixing it and move on.}
\end{quote}

\textbf{Trigger Query (Weeks Later):}
\begin{quote}
\textit{A: I redesigned the workflow to make sure the same mistake can’t
happen again and asked two coworkers to double-check it.}
\end{quote}

\textbf{Analysis.}
The later utterance does not restate the emotional state.
Correct handling requires remembering the earlier distress and inferring
that lingering anxiety motivated cautious, preventive behavior.
This type of memory cannot be evaluated by checking for factual overlap,
as the relevant signal lies in the persistence of an internal state.

\paragraph{Example 3: Goal-Oriented Memory (Long-Term Intention and Re-evaluation).}

\textbf{Cue Dialogue (Goal):}
\begin{quote}
\textit{A: I’m saving up specifically for a vintage convertible because
I’ve always dreamed of owning a classic car.} \\
\textit{B: That sounds like a stylish goal.}
\end{quote}

\textbf{Trigger Query (Several Months Later):}
\begin{quote}
\textit{A: Lately I’ve realized my happiest moments come from camping
trips, not from buying things.}
\end{quote}

\textbf{Analysis.}
This example probes whether the model can recognize a potential shift or
re-evaluation of a previously stated long-term goal.
There is no factual question to answer and no explicit contradiction in
surface form.
Instead, the model must reason about how a new reflection relates to an
earlier intention, highlighting the dynamic nature of goal memory.

\paragraph{Example 4: Value-Based Memory (Consistency Across Contexts).}

\textbf{Cue Dialogue (Value):}
\begin{quote}
\textit{A: I turn down clients with unrealistic timelines because I
value my team’s well-being more than profit.} \\
\textit{B: That says a lot about the kind of leader you are.}
\end{quote}

\textbf{Trigger Query (Later):}
\begin{quote}
\textit{A: I watched my kid fall asleep over homework and suddenly
realized I’ve protected my team from burnout for years while ignoring
how exhausted my own family looks.}
\end{quote}

\textbf{Analysis.}
No explicit fact is queried in the trigger.
Correct interpretation requires recalling the earlier value statement
and recognizing a value-consistency tension across different social
roles.
Such cases emphasize reasoning over personal principles rather than
retrieval of explicit information.

\paragraph{Summary.}
These examples demonstrate that cognitive memory cases in our dataset
differ fundamentally from fact-based benchmarks.
Correctness is determined by whether a response reflects awareness and
appropriate use of previously expressed constraints, rather than by
reproducing a specific reference string.
Because cognitive memory constitutes a core contribution of this work,
we include only a subset of representative examples in the paper and
appendix.
The complete set of cognitive instances, along with full annotations and
LLM-based judge rationales, is released in our code repository for
comprehensive inspection and reproducibility.

\begin{table*}[t]
\centering
\small
\renewcommand{\arraystretch}{1.5}
\begin{tabularx}{\textwidth}{l | X}
\toprule
\textbf{Annotation} & \textbf{Full Prompt Content} \\ \midrule

\begin{tabular}[t]{@{}l@{}}
    \textcolor{cRole}{$\bullet$ Role Setting} \\
    \textcolor{cLogic}{$\bullet$ Task Logic} \\
    \textcolor{cReq}{$\bullet$ Constraints} \\
    \textcolor{cFormat}{$\bullet$ Output Format}
\end{tabular} & 
\textcolor{cRole}{You are generating short conversational cues for testing memory recall capabilities. Generate short dialogues (2 lines) based on the relation type provided. Relation Type: \{relation\_type\}} 

\textcolor{cLogic}{Relation type meanings: - Causal: an earlier cause or condition affects a later event. - State: a physical or emotional state influences later behavior. - Goal: a long-term intention or plan influences current choices. - Value: a belief or value shapes later reactions.}

\textcolor{cReq}{Requirements: - Create exactly 2 lines of dialogue for each example: (A) Mentions a MEMORABLE and RECALLABLE event; (B) Gives a short, natural reaction that CLOSES the conversation. - Ensure dialogue closure. - Memory anchor: A's line should contain a distinctive detail. - Make it sound like realistic, natural daily conversation. - Vary topics across work, family, relationships, health, travel, etc.}

\textcolor{cFormat}{- Do NOT include explanations or markdown. Output a valid JSON array ONLY. Output strictly in this format: [ \{ "relation\_type": "...", "cue\_dialogue": "A: ...\textbackslash nB: ..." \} ]. Generate \{num\_samples\} examples.} \\ 

\bottomrule
\end{tabularx}
\caption{Cue generation prompt with semantic annotations.}
\label{tab:cue_prompt}
\end{table*}

\begin{table*}[t]
\centering
\small
\renewcommand{\arraystretch}{1.5}
\begin{tabularx}{\textwidth}{l | X}
\toprule
\textbf{Annotation} & \textbf{Full Prompt Content} \\ \midrule

\begin{tabular}[t]{@{}l@{}}
    \textcolor{cRole}{$\bullet$ Role Setting} \\
    \textcolor{cLogic}{$\bullet$ Task Logic} \\
    \textcolor{cReq}{$\bullet$ Constraints} \\
    \textcolor{cFormat}{$\bullet$ Output Format}
\end{tabular} & 
\textcolor{cRole}{You are generating trigger queries that have implicit cognitive connections to given dialogues and create meaningful cognitive conflicts or contrasts with given dialogues, ensuring diverse perspectives in memory recall.} 

\vspace{0.5em}
\textcolor{cLogic}{CRITICAL REQUIREMENT: Each of the five trigger queries must represent a DISTINCT COGNITIVE ANGLE of conflict or recall. They should not feel similar or repetitive. Aim for five truly different ways that the trigger could relate to the cue. Given the cue dialogue below, generate FIVE DIFFERENT trigger queries that: 1. Have implicit connections (memory recall); 2. Are semantically distant; 3. Sound natural; 4. Represent events one week or months after; 5. The time gaps should be at least one week or more; 6. Create some conflict or contrast.} 

\vspace{0.5em}
\textcolor{cReq}{Each trigger query should: - Be spoken by the same person (A) in first person; - Sound semantically unrelated but cognitively connected; - Be something humans can recall from, but similarity-based retrievers cannot; - Be a statement, feeling, question, or reflection; - Avoid reusing nouns or verbs from the cue; - Represent a distinct angle from others. AVOID: Similar-sounding triggers, superficial connections, forced contrasts, or repeating relationship types.}

\vspace{0.5em}
\textcolor{cFormat}{Requirements: Generate FIVE DISTINCT trigger queries; Vary topics and time gaps (one week to several months); Do NOT include explanations; Include time\_gap description. Cue Dialogue: \{cue\_dialogue\} Relation Type: \{relation\_type\} Output strictly in this format: [ \{ "relation\_type": "\{relation\_type\}", "cue\_dialogue": "\{cue\_dialogue\}", "trigger\_query": "A: ...", "time\_gap": "..." \}, ... ]} \\ 

\bottomrule
\end{tabularx}
\caption{Trigger query generation prompt with semantic annotations.}
\label{tab:trigger-prompt}
\end{table*}

\begin{table*}[ht]
\centering
\small
\renewcommand{\arraystretch}{1.5}
\begin{tabularx}{\textwidth}{l | X}
\toprule
\textbf{Annotation} & \textbf{Full Prompt Content (Evaluation Templates)} \\ \midrule

\textbf{Single/Multi-hop/Commonsense} & \\
\midrule
\textcolor{cRole}{$\bullet$ Role Setting} & \textcolor{cRole}{You are a Fact-Checking or Commonsense Judge. Your task is to compare the prediction with the reference answer using external knowledge where needed.} \\
\textcolor{cLogic}{$\bullet$ Task Logic} & \textcolor{cLogic}{Question: \{question\} Reference Answer: \{gold\} Model Prediction: \{pred\} Relevant Evidence: \{evidence\}} \\
\textcolor{cConstraints}{$\bullet$ 3-Level Labels} & \textcolor{cConstraints}{Labels: - "correct": exact match or sound inference; - "partial": minor inaccuracies or incomplete reasoning; - "wrong": factually incorrect or contradicts commonsense.} \\
\textcolor{cFormat}{$\bullet$ Format} & \textcolor{cFormat}{Return your judgment strictly in JSON format: \{"label": "...", "reason": "..."\}} \\ \midrule

\textbf{Temporal Reasoning} & \\
\midrule
\textcolor{cRole}{$\bullet$ Role Setting} & \textcolor{cRole}{You are a Temporal Logic Judge. Your task: Check the calculation, duration, or sequence of events strictly.} \\
\textcolor{cLogic}{$\bullet$ Task Logic} & \textcolor{cLogic}{Question: \{question\} Reference Answer: \{gold\} Model Prediction: \{pred\}} \\
\textcolor{cConstraints}{$\bullet$ Binary Labels} & \textcolor{cConstraints}{Labels: - "correct": calculated time or sequence matches exactly; - "wrong": calculation is incorrect or sequence is reversed. **Note**: Precision is key, no partial credit.} \\
\textcolor{cFormat}{$\bullet$ Format} & \textcolor{cFormat}{Return your judgment strictly in JSON format: \{"label": "...", "reason": "..."\}} \\ \midrule

\textbf{Adversarial Robustness} & \\
\midrule
\textcolor{cRole}{$\bullet$ Role Setting} & \textcolor{cRole}{You are a Skeptical Judge. Determine if the model correctly identifies unanswerable questions or semantic conflicts.} \\
\textcolor{cLogic}{$\bullet$ Task Logic} & \textcolor{cLogic}{Question: \{question\} Reference Answer: \{gold\} Model Prediction: \{pred\} Relevant Evidence: \{evidence\}} \\
\textcolor{cConstraints}{$\bullet$ Refusal Check} & \textcolor{cConstraints}{Labels: - "correct": model correctly refuses to answer or identifies the non-existent event; - "wrong": model hallucinates an answer or provides incorrect info.} \\
\textcolor{cFormat}{$\bullet$ Format} & \textcolor{cFormat}{Return your judgment strictly in JSON format: \{"label": "...", "reason": "..."\}} \\ \midrule

\textbf{Cognitive Awareness} & \\
\midrule
\textcolor{cRole}{$\bullet$ Role Setting} & \textcolor{cRole}{You are a Memory Awareness Judge. Determine if the model prediction demonstrates awareness of the memory cue found in the Evidence.} \\
\textcolor{cLogic}{$\bullet$ Scenario} & \textcolor{cLogic}{Scenario: Evidence contains a specific user memory; Question is a trigger interacting with that memory.} \\
\textcolor{cConstraints}{$\bullet$ Recall Check} & \textcolor{cConstraints}{Labels: - "correct": explicitly acknowledges or adapts to the Memory/Cue (proves recall); - "wrong": completely ignores the Evidence and gives a generic response.} \\
\textcolor{cFormat}{$\bullet$ Format} & \textcolor{cFormat}{Return your judgment strictly in JSON format: \{"label": "...", "reason": "..."\}} \\ \bottomrule

\end{tabularx}
\caption{Comprehensive evaluation prompt templates across four dimensions: Fact/Commonsense, Temporal, Adversarial, and Cognitive.}
\label{tab:unified-evaluation-prompts}
\end{table*}

\end{document}